\documentclass{article}
\usepackage{ijcai19}

\usepackage{times}
\usepackage{soul}
\usepackage{url}
\usepackage[hidelinks]{hyperref}
\usepackage[utf8]{inputenc}
\usepackage[small]{caption}
\usepackage{graphicx}
\usepackage{subcaption}
\usepackage{comment}
\usepackage{amsmath}
\usepackage{booktabs}
\usepackage{algorithm}
\usepackage{algorithmic}
\usepackage{footnote}
\renewcommand{\thefootnote}{\alph{footnote}}

\newcommand{\astfootnote}[1]{
\let\oldthefootnote=\thefootnote
\setcounter{footnote}{0}
\renewcommand{\thefootnote}{\fnsymbol{footnote}}
\footnote{#1}
\let\thefootnote=\oldthefootnote
}

\newcommand\blfootnote[1]{%
  \begingroup
  \renewcommand\thefootnote{}\footnote{#1}%
  \addtocounter{footnote}{-1}%
  \endgroup
}

\title{Ex-Twit: Explainable Twitter Mining on Health Data}

\author{
    Tunazzina Islam
    \affiliations
    Department of Computer Science, \\
    Purdue University, West Lafayette, \\
    Indiana 47907, USA. \emails
    islam32@purdue.edu
}

\begin{document}

\maketitle

\begin{abstract}
Since most machine learning models provide no explanations for the predictions, their predictions are obscure for the human. The ability to explain a model's prediction has become a necessity in many applications including Twitter mining. In this work, we propose a method called Explainable Twitter Mining (Ex-Twit) combining Topic Modeling and Local Interpretable Model-agnostic Explanation (LIME) to predict the topic and explain the model predictions. We demonstrate the effectiveness of Ex-Twit on Twitter health-related data.
\end{abstract}

\blfootnote{In Proceedings of 7$^{th}$ International Workshop on  Natural Language Processing for Social Media (SocialNLP 2019) at IJCAI 2019 in Macao, China.}

\section{Introduction}
With the increasing availability and efficiency of modern computing, machine learning systems have become increasingly powerful and complex. Dramatic success in machine learning has led to an explosion of AI applications \cite{gunning2017explainable}. The problem with most of the machine learning algorithms is that they are designed to be trained over training examples and their predictions are obscure for the human. There is a vital concern regarding trust. Explainable AI (XAI) provides details explanations of the decisions in some level \cite{gilpin2018explaining}. These explanations are important to ensure fairness of machine learning algorithm and identify potential bias in the training data.

The main motivation of this work has been started with few questions (1) \textit{What do people do to maintain their health?}-- some people do balanced diet, some do exercise. Among diet plans some people maintain vegetarian diet/vegan diet, among exercises some people do swimming, cycling or yoga. Now-a-days people usually share their interest, thoughts via discussions, tweets, status in social media (i.e. Facebook, Twitter, Instagram etc.). It's huge amount of data and it's not possible to go through all the data manually. This motivates us to do Twitter mining on health-related data using Topic Modeling. (2) \textit{To what extend we could trust the model predictions?} This research question is one of the main motivations of our work to explain the prediction of model. 

Twitter has been growing in popularity and now-a-days, it is used everyday by people to express opinions about different topics, such as products, movies, health, music, politicians, events, among others. Twitter data constitutes a rich source that can be used for capturing information about any topic imaginable. This data can be used in different use cases such as finding trends related to a specific keyword, measuring brand sentiment, and gathering feedback about new products and services. In this work, we use text mining to mine the Twitter health-related data. Text mining is the application of natural language processing techniques to derive relevant information \cite{allahyari2017brief}. This is getting a lot attention these last years, due to an exponential increase in digital text data from web pages. 
\begin{figure}[htbp]
  \centering  
  \includegraphics[width= 0.5\textwidth]{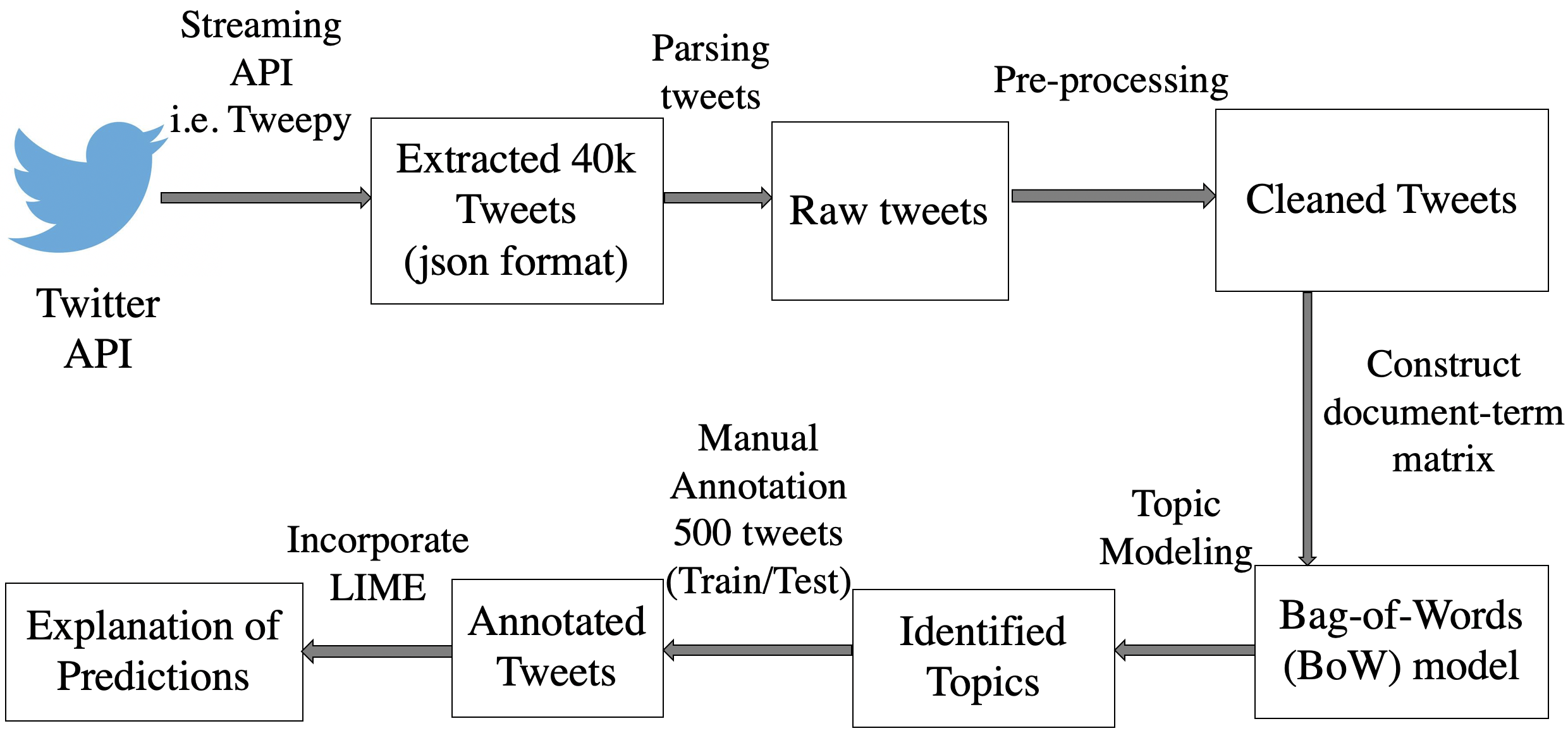}
    \caption{Methodology of Ex-Twit: Explainable Twitter mining on health data.}
    \label{fig:method}
\end{figure}
\begin{figure*}
\centering
    \begin{subfigure}[b]{0.4\textwidth}            
            \includegraphics[width=\textwidth]{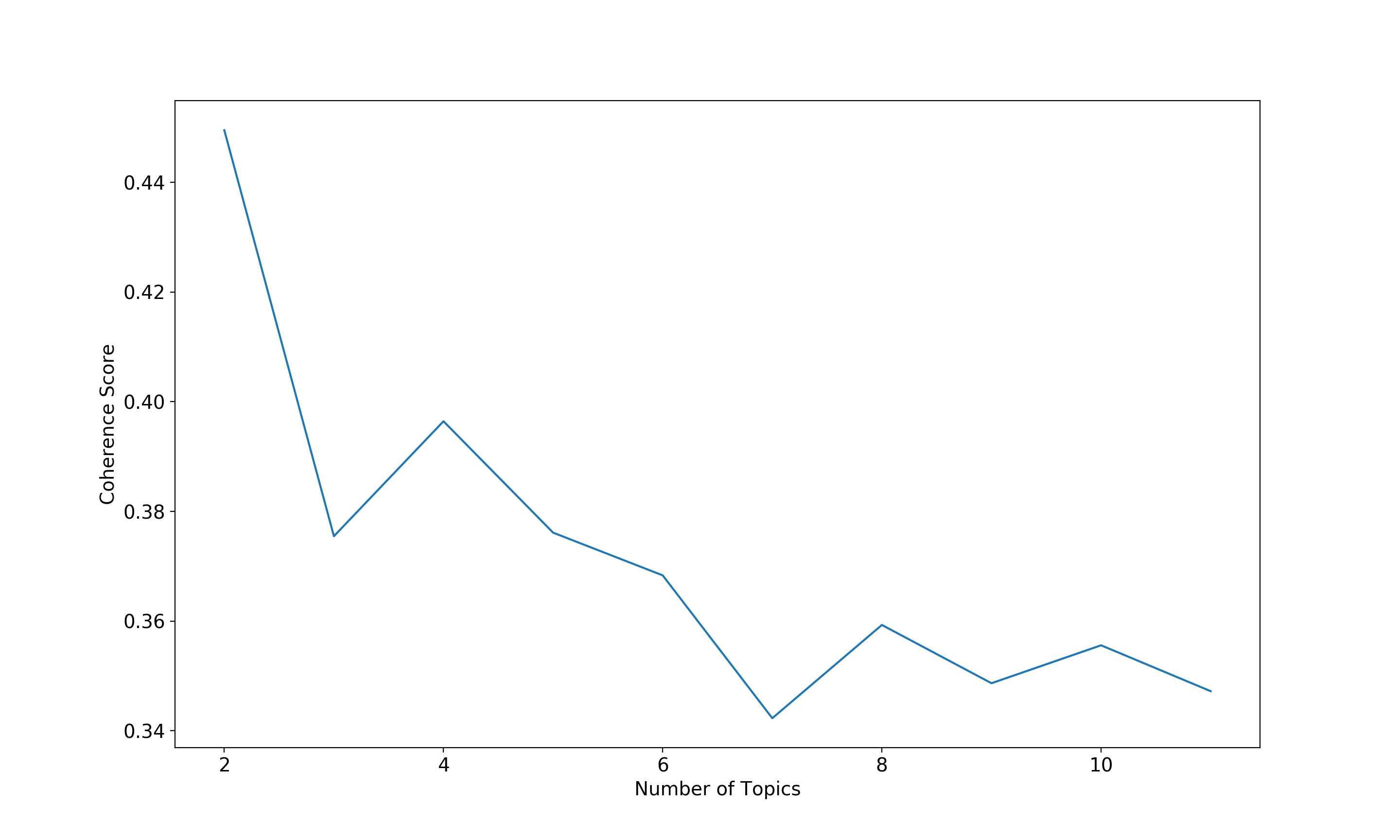}
            \caption{Number of Topics = 2 in LSA.}
            \label{fig:LSA}
    \end{subfigure}
    \begin{subfigure}[b]{0.4\textwidth}
            \centering
            \includegraphics[width=\textwidth]{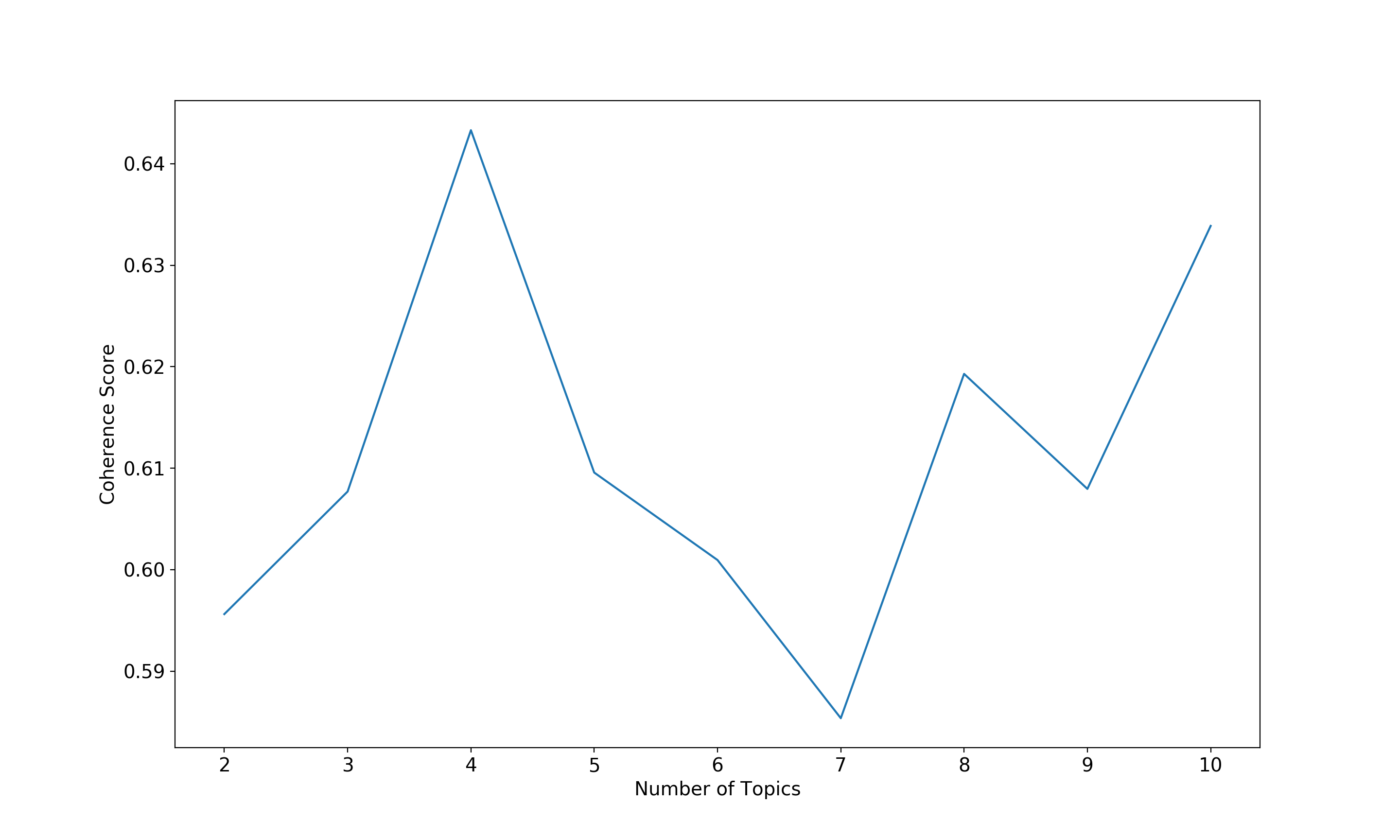}
            \caption{Number of Topics = 4 in NMF.}
            \label{fig:NMF}
    \end{subfigure}
    \begin{subfigure}[b]{0.4\textwidth}
            \centering
            \includegraphics[width=\textwidth]{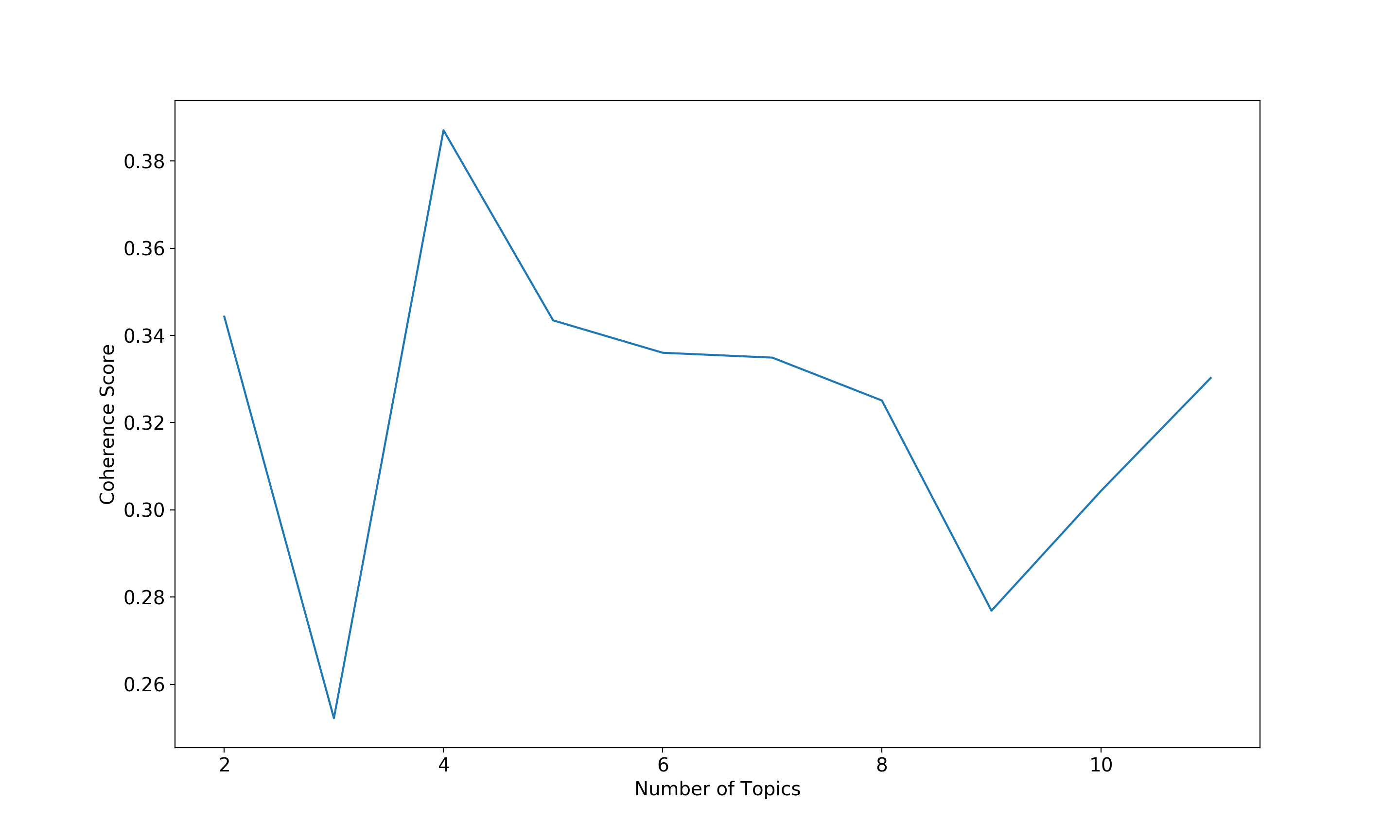}
            \caption{Number of Topics = 4 in LDA.}
            \label{fig:LDA}
    \end{subfigure}
    
    \caption{Optimal Number of Topics vs Coherence Score. Number of Topics (k) are selected based on the highest coherence score. Graphs are rendered in high resolution and can be zoomed in.}\label{fig:k_coherence}
\end{figure*}
In this paper, we use topic modeling to infer semantic structure of the unstructured data (i.e Tweets). Topic Modeling is a text mining technique which automatically discovers the hidden themes from given documents. It is an unsupervised text analytic algorithm that is used for finding the group of words from the given document. In our work, Topic model predicts the topics of corresponding tweets. In order to trust black-box methods, we need explainability. We incorporate Local Interpretable Model-agnostic Explanation (LIME) \cite{ribeiro2016should} method with topic modeling for the explanation of model predictions. We build a pipeline called Ex-Twit: Explainable Twitter Mining on Health Data to explain the prediction of topic modeling. Finally, we conduct experiment on real Twitter data and show that our approach successfully predicts the topic with explanations.

Fig. \ref{fig:method} shows the overall pipeline of Ex-Twit. The implication of our work is to explain the annotation of unlabeled data done by the model. 

\section{Related Work}

Use of data generated through social media is gradually increasing. Several works have been done on prediction of social media content \cite{son2017recognizing}; \cite{yaden2018language}; \cite{eichstaedt2018facebook}; \cite{de2013predicting}; \cite{bian2012towards}; \cite{pandarachalil2015twitter}; \cite{pak2010twitter}; \cite{cobb2012health}. Social media support the collection and analysis of data in real time in the real world. 

Twitter has been growing in popularity now-a-days. \cite{prieto2014twitter} proposed a method to extract a set of tweets to estimate and track the incidence of health conditions in society. Discovering public health topics and themes in tweets had been examined by \cite{prier2011identifying}. \cite{yoon2013practical} described a practical approach of content mining to analyze tweet contents and illustrate an application of the approach to the topic of physical activity. \cite{reece2017forecasting} developed computational models to predict the emergence of depression and Post-Traumatic Stress Disorder (PTSD) in Twitter users. \cite{wang2017twiinsight} developed a system called TwiInsight to identify the insight from Twitter data. \cite{islam2019yoga} did correlation mining to explore hidden patterns and unknown correlations in Twitter health-related data. In our work, we use Topic Modeling to extract topic from health-related tweets.

Machine learning systems have become increasingly powerful and complex day by day. Developing sparse interpretable models is of considerable interest to the broader research community \cite{letham2015interpretable}; \cite{kim2015mind}; \cite{lei2016rationalizing}. To justify predictions made by neural networks, \cite{lei2016rationalizing} built modular neural framework to automatically generate concise yet sufficient text
fragments. \cite{ribeiro2016should} proposed a model agnostic framework (LIME) where the proxy model is learned ensuring locally valid approximations for the target sample and its neighborhood.  In this paper, we incorporate LIME \cite{ribeiro2016should} framework with topic modeling.

To justify the model’s prediction on Twitter mining, we propose a method combining topic modeling  and  LIME  model  explanation named Ex-Twit to predict the health-related topic and explain the model predictions for unsupervised learning. 

\begin{table*}[htbp]
  \centering
  \caption{Topics and top-10 keywords of the corresponding topic}
    \resizebox{\textwidth}{!}{\begin{tabular}{|c|c||c|c|c|c||c|c|c|c|}
    \hline
    \multicolumn{2}{|c||}{\textbf{LSA}} & \multicolumn{4}{|c||}{\textbf{NMF}} & \multicolumn{4}{|c|}{\textbf{LDA}} \\
    \hline
    \multicolumn{1}{|c|}{\textit{Topic 1}} & \multicolumn{1}{|c||}{\textit{Topic 2}} & \multicolumn{1}{|c|}{\textit{Topic 1}}& \multicolumn{1}{|c|}{\textit{Topic 2}} & \multicolumn{1}{|c|}{\textit{Topic 3}} & \multicolumn{1}{|c||}{\textit{Topic 4}} & \multicolumn{1}{|c|}{\textit{Topic 1}} & \multicolumn{1}{|c|}{\textit{Topic 2}} & \multicolumn{1}{|c|}{\textit{Topic 3}} & \multicolumn{1}{|c|}{\textit{Topic 4}} \\
    \hline
    Yoga  & diet  & Yoga  & diet  & vegan & fitness & diet  & vegan  & swimming  & fitness  \\
    \hline
    everi & vegan & job   & beyonce & go & workout & workout & yoga  & swim  & amp \\
    \hline
    Life  & fit   & every\_woman & new   & eat   & go    & new   & job   & day   & wellness \\
    \hline
    Job   & day   & cooks\_goe & bitch & make  & good  & go    & every\_woman   & much  & health  \\
    \hline
    Remember & new   & therapy\_remember & ciara\_prayer & food  & amp   & day   & cooks\_goe  & support  & time  \\
    \hline
    goe   & like  & life\_juggl & day   & day   & day   & beyonce & therapy\_remember &  really  & great  \\
    \hline
    Woman & Beyonce & everyone\_birthday & eat   & amp   & yoga  & get   & life\_juggle  & try   & look \\
    \hline
    Everyone & amp   & boyfriend & go    & shit  & health & today  & everyone\_birthday & always & hiking  \\
    \hline
    cook  & eat   & hot   & fat   & meat  & gym   & bitch & eat   & relationship & make \\
    \hline
    therapy & workout & know  & keto  & vegetarian & today & gym   & boyfriend & pool  & love \\
    \hline
    \end{tabular}}
  \label{tab:topic_keywords}%
\end{table*}%

\section{Data Collection}
We use Twitter health-related data for Ex-Twit. In \hyperref[subsec:3.1]{Subsection 3.1}, we show data crawling process from Twitter. Subsection \hyperref[subsec:3.2]{3.2} shows data pre-processing method.
\subsection{Data Extraction}
\label{subsec:3.1}
The twitter data has been crawled using Tweepy which is a Python library 
for accessing the Twitter API. We use Twitter streaming API to extract 40k tweets (April 17-19, 2019). For the crawling, we focus on several keywords that are related to health. The keywords are processed in a non-case-sensitive way. We use filter to stream all tweets containing the word `yoga', `healthylife', `healthydiet', `diet',`hiking', `swimming', `cycling', `yogi', `fatburn', `weightloss', `pilates', `zumba', `nutritiousfood', `wellness', `fitness', `workout', `vegetarian', `vegan', `lowcarb', `glutenfree', `calorieburn'. 

\subsection{Data Pre-processing}
\label{subsec:3.2}
Data pre-processing is one of the key components in many text mining
algorithms \cite{allahyari2017brief}. Data cleaning is crucial for generating a useful topic model. We have some prerequisites i.e. we download the stopwords from NLTK (Natural Language Toolkit) and spacy's en model for text pre-processing. 

It is noticeable that the parsed full-text tweets have many emails, `RT', newline and extra spaces that is quite distracting. We use Python Regular Expressions (re module) to get rid of them.

After removing the emails and extra spaces, we tokenize each text into a list of words, remove punctuation and unnecessary characters. We use Python Gensim package for further processing. Gensim's simple\_preprocess() is used for tokenization and removing punctuation. We use Gensim's Phrases model to build bigrams. Certain parts of English speech, like conjunctions (``for", ``or") or the word ``the" are meaningless to a topic model. These terms are called stopwords and we remove them from the token list. We use spacy model for lemmatization to keep only noun, adjective, verb, adverb. Stemming words is another common NLP technique to reduce topically similar words to their root. For example, ``connect", ``connecting", ``connected", ``connection", ``connections" all have similar meanings; stemming reduces those terms to “connect”. The Porter stemming algorithm \cite{porter1980algorithm} is the most widely used method.

\section{Methodology}
In this paper, we show a method to explain unsupervised learning. Our approach called Ex-Twit combining topic modeling and LIME model explanation predicts and explains health-related topics extracted from Twitter. Subsections \hyperref[subsec:4.1]{4.1} and \hyperref[subsec:4.2]{4.2} elaborately present how we can infer the meaning of unstructured data. Subsection \hyperref[subsec:4.3]{4.3} shows how we do manual annotation for ground truth comparison. We show the method for explainability in \hyperref[subsec:4.4]{Subsectuon 4.4}.  Fig. \ref{fig:method} shows the methodology of Ex-Twit.

\subsection{Construct bag-of-words Model}
\label{subsec:4.1}
The result of the data cleaning stage is texts, a tokenized, stopped, stemmed and lemmatized list of words from a single tweet. To understand how frequently each term occurs within each tweet, we construct a document-term matrix using Gensim's Dictionary() function. Gensim's doc2bow() function converts dictionary into a bag-of-words. In the bag-of-words model, each tweet is represented by a vector in a m-dimensional coordinate space, where m is number of unique terms across all tweets. This set of terms is called the corpus vocabulary.


\subsection{Choose optimal number of Topics}
\label{subsec:4.2}
Topic modeling is a text mining technique which provides methods for identifying co-occurring keywords to summarize collections of textual information. This is used to analyze collections of documents, each of which is represented as a mixture of topics, where each topic is a probability distribution over words \cite{alghamdi2015survey}. Applying these models to a document collection involves estimating the topic distributions and the weight each topic receives in each document. A number of algorithms exist for solving this problem. We use three unsupervised machine learning algorithms to explore the topics of the tweets: Latent Semantic Analysis (LSA) \cite{deerwester1990indexing}, Non-negative Matrix Factorization (NMF) \cite{lee2001algorithms}, and Latent Dirichlet Allocation (LDA) \cite{blei2003latent}. 

Topic modeling is an unsupervised learning, so the set of possible topics are unknown. To find out the optimal number of topic, we build many LSA, NMF, LDA models with different values of number of topics (k) and pick the one that gives the highest coherence score. Choosing a `k' that marks the end of a rapid growth of topic coherence usually offers meaningful and interpretable topics. 

We use Gensim's coherencemodel to calculate topic coherence for topic models (LSA and LDA). For NMF, we use a topic coherence measure called TC-W2V. This measure relies on the use of a word embedding model constructed from the corpus. So in this step, we use the Gensim implementation of Word2Vec \cite{mikolov2013efficient} to build a Word2Vec model based on the collection of tweets.

We achieve the highest coherence score = 0.4495 when the number of topics is 2 for LSA, for NMF the highest coherence value is 0.6433 for K = 4, and for LDA we also get number of topics is 4 with the highest coherence score which is 0.3871 (see Fig. \ref{fig:k_coherence}). 

For our dataset, we picked k = 2, 4, and 4 with the highest coherence value for LSA, NMF, and LDA correspondingly (Fig. \ref{fig:k_coherence}). Table \ref{tab:topic_keywords} shows the topics and top-10 keywords of the corresponding topic. We get more informative and understandable topics using LDA model than LSA. LSA decomposed matrix is a highly dense matrix, so it is difficult to index individual dimension. LSA unable to capture the multiple meanings of words. It offers lower accuracy than LDA.

In case of NMF, we observe same keywords are repeated in multiple topics. Keywords ``go", ``day" both are repeated in Topic 2, Topic 3, and Topic 4 (See 4$^{th}$, 5$^{th}$, and 6$^{th}$ columns of Table \ref{tab:topic_keywords}). In Table \ref{tab:topic_keywords} keyword ``yoga" has been found both in Topic 1 (3$^{rd}$ column) and Topic 4 (6$^{th}$ column). We also notice that keyword ``eat" is in Topic 2 and Topic 3 (Table \ref{tab:topic_keywords} 4$^{th}$ and 5$^{th}$ columns). If the same keywords being repeated in multiple topics, it is probably a sign that the `k' is large though we achieve the highest coherence score in NMF for k=4.

We choose topics generated by LDA model for our further analysis. Because LDA is good in identifying coherent topics where as NMF usually gives incoherent topics. However, in the average case NMF and LDA are similar but LDA is more consistent. 

\subsection{Manual Annotation}
\label{subsec:4.3}
To calculate the accuracy of model in comparison with ground truth label, we selected top 500 tweets from train dataset (40k tweets). We extracted 500 new tweets (22 April, 2019) as a test dataset. We did manual annotation both for train and test data by choosing one topic among the 4 topics generated from LDA model (7$^{th}$, 8$^{th}$, 9$^{th}$, and 10$^{th}$ columns of Table \ref{tab:topic_keywords}) for each tweet based on the intent of the tweet. Consider the following two tweets:

Tweet 1: \textit{Learning some traditional yoga with my good friend.}

Tweet 2: \textit{Why You Should \#LiftWeights to Lose \#BellyFat \#Fitness \#core \#abs \#diet \#gym \#bodybuilding \#workout \#yoga}

The intention of Tweet 1 is yoga activity (i.e. learning yoga). Tweet 2 is more about weight lifting to reduce belly fat. This tweet is related to workout. When we do manual annotation, we assign Topic 2 in Tweet 1, and Topic 1 in Tweet 2. It's not wise to assign Topic 2 for both tweets based on the keyword ``yoga". We assign zero if we could not infer which topic goes well with the corresponding tweet. During annotation, we focus on functionality of tweets.
\begin{table*}[htbp]
  \centering
  \caption{Explanation Observation}
    \resizebox{\textwidth}{!}{\begin{tabular}{|c|p{22.585em}|c|c|c|c|}
    \hline
    \multicolumn{1}{|c|}{\textbf{Dataset}} & \multicolumn{1}{|c|}{\textbf{Tweets}} & \multicolumn{1}{c|}{\textbf{Predicted Topic}} & \multicolumn{1}{c|}{\textbf{Manual Annotation}} & \multicolumn{1}{c|}{\textbf{mean KL Divergence}} & \multicolumn{1}{c|}{\textbf{Score}} \\
    \hline
     Train & Revoking my vegetarian status till further notice. There's something I wanna do and I can't afford the supplements that come with being veggie. & 2     & 2     & 0.026 & 1.0 \\ 
    \hline
    Train & Burned 873 calories doing 52 minutes of Swimming laps, freestyle, light/moderate effort \#myfitnesspal \#myfitnesspal. & 3     & 3     & 0.015 & 0.95 \\
    \hline
    Test  & This morning I packed myself a salad. Went to yoga during lunch. And then ate my salad with water in hand. I'm feeling so healthy I don't know what to even do with myself. Like maybe I should eat a bag of chips or something. & 2     & 2     & 0.024 & 0.95 \\
    \hline
     Test  & I lost 28kg / 61 lbs in 6 months! I changed my diet and went gym 5/6 times a week. I'm lighter and much happier now. & 1     & 1     & 0.025 & 0.94 \\
    \hline
     Test  & My weight two days ago was 149.7. I almost cried. Two years ago I whittled my way down from around 160 to around 145. It felt So good. In the past year my weight has been as low as 142. If only it were lower than that now, \#Journal \#WeightLoss. & 1     & 1     &  0.03 & 0.97 \\
    \hline
     Test  &  Swimming is great. It's a perfect workout. \#fitness \#wellness & 4     & 3     & 0.061 & 0.92 \\
    \hline
    \end{tabular}}
  \label{tab:observ}
\end{table*} 
\subsection{Prediction Explanation}
\label{subsec:4.4}
It's easier for human to infer the topic from the tweets and human can explain the reason of the inference. But machine learning models remain mostly black boxes. Explainable machine learning model can present textual or visual artifacts that provide qualitative understanding of the
relationship between the instance's components (e.g. words
in text) and the model's prediction \cite{ribeiro2016should}. We incorporate LIME \cite{ribeiro2016should} method for the explanation of model predictions of twitter mining for health data. 
\begin{figure}
\centering
    \begin{subfigure}[b]{0.5\textwidth}            
            \includegraphics[width=\textwidth]{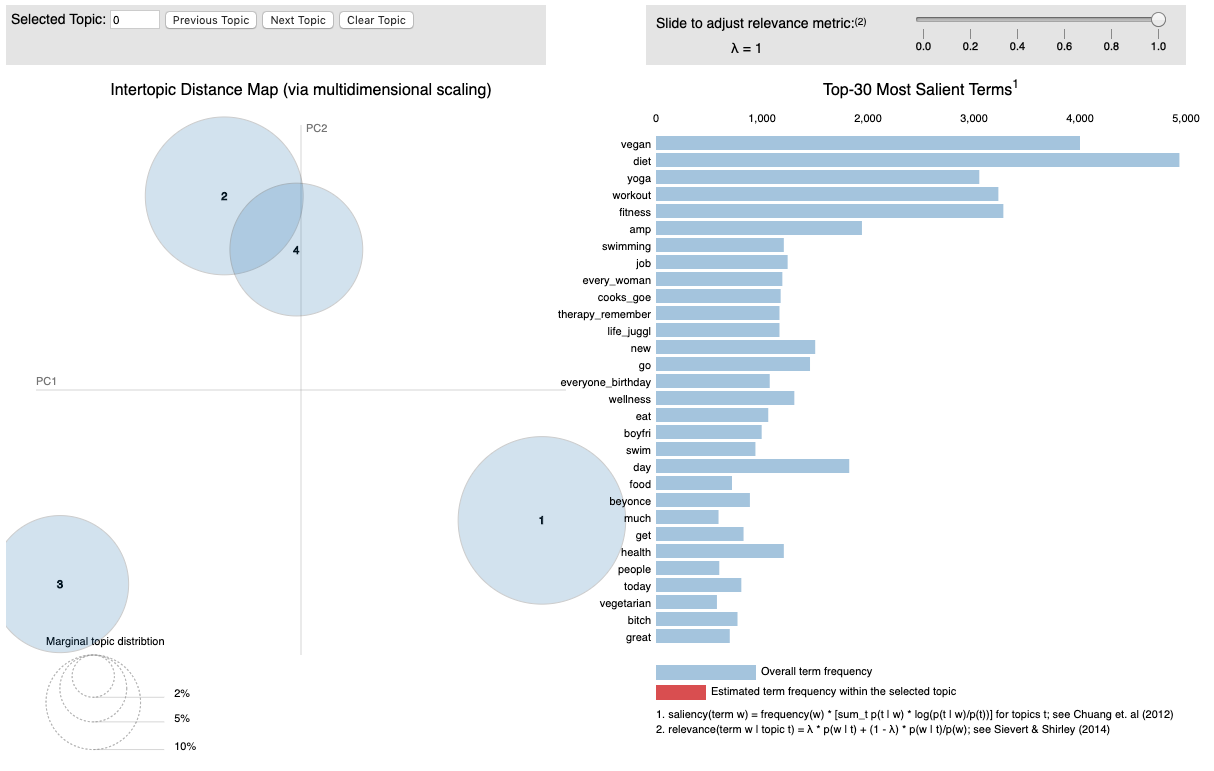}
            \caption{Bubbles in left hand side show overall topic distribution and sky blue bars in right hand side represent overall term frequencies. Best viewed in electronic format (zoomed in).}
            \label{fig:LDAVis1}
    \end{subfigure}
    \begin{subfigure}[b]{0.5\textwidth}
            \centering
            \includegraphics[width=\textwidth]{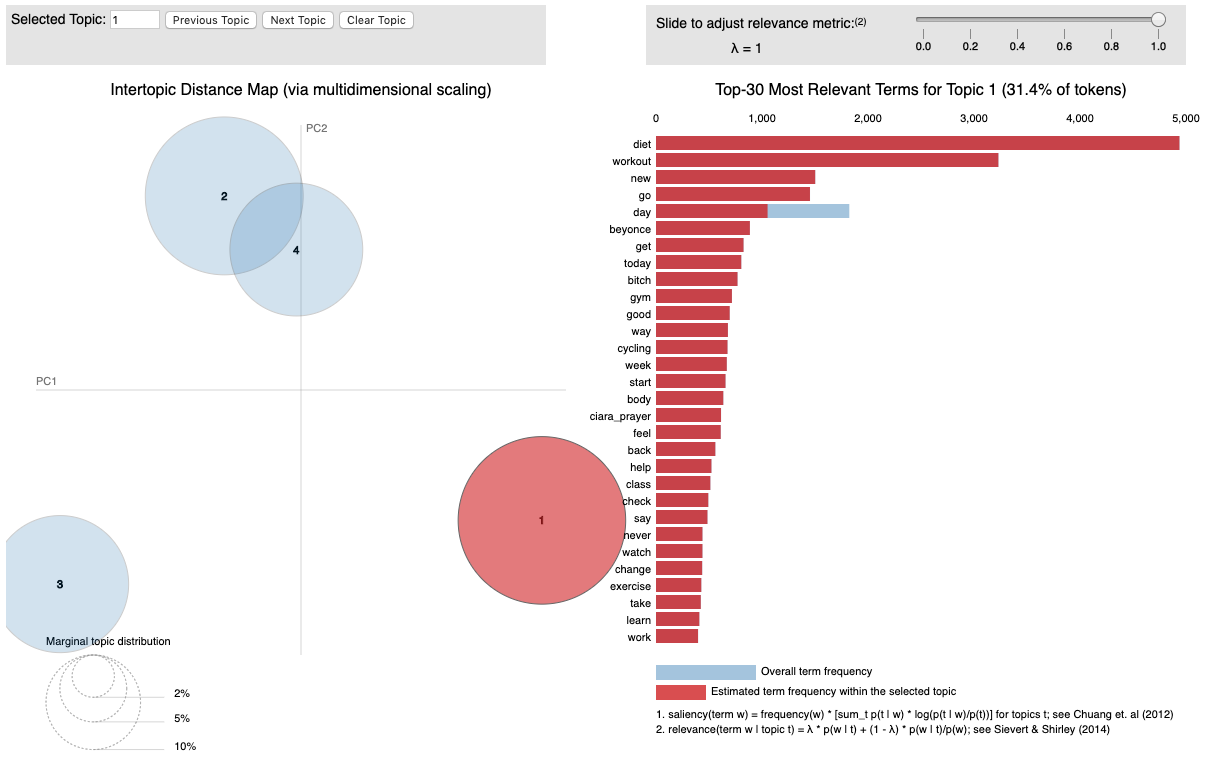}
            \caption{Red bubble in left hand side represents selected Topic which is Topic 1. Red bars in right hand side show estimated term frequencies of top-30 salient keywords that form the  Topic 1. Best viewed in electronic format (zoomed in).}
            \label{fig:LDAVis2}
    \end{subfigure}
    
    \caption{Visualization using pyLDAVis. Best viewed in electronic format (zoomed in).}\label{fig:LDAVis}
\end{figure}

The LIME algorithm generates distorted versions of the tweets by removing some of the words from the tweets; predict probabilities for these distorted tweets using the black-box classifier; train another classifier which tries to predict output of a black-box classifier on these tweets.

We evaluate the explanation by observing the accuracy score weighted by cosine distance between generated sample and original tweets and Kullback–Leibler divergence (KL divergence) which is also weighted by distance. KL divergence 0.0 means a perfect match.

\section{Results and Discussion}
\subsection{Visualization}
We use LDAvis \cite{sievert2014ldavis}, a web-based interactive visualization of topics estimated using LDA. Gensim's pyLDAVis is the most commonly used visualization tool to visualize the information contained in a topic model. In Fig. \ref{fig:LDAVis}, each bubble on the left-hand side plot represents a topic. The larger the bubble, the more prevalent is that topic. A good topic model has fairly big, non-overlapping bubbles scattered throughout the chart instead of being clustered in one quadrant. A model with too many topics, is typically have many overlaps, small sized bubbles clustered in one region of the chart. In right hand side, the words represent the salient keywords.

If we move the cursor over one of the bubbles (Fig. \ref{fig:LDAVis2}), the words and bars on the right-hand side have been updated and top-30 salient keywords that form the selected topic and their estimated term frequencies are shown.
\begin{figure}[htbp]
  \centering  
  \includegraphics[width= 0.5\textwidth]{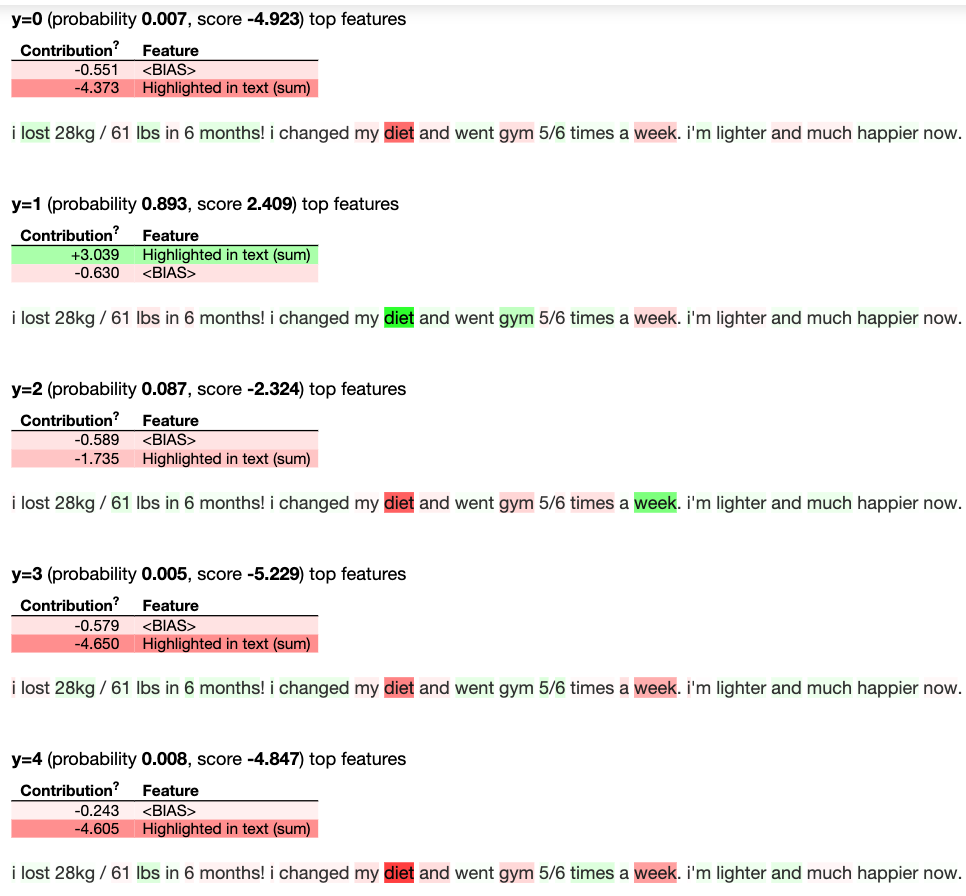}
    \caption{Explainable output of model prediction using Ex-Twit where human inference matches with model prediction. Here, y represents topic number. y = 0 means incomprehensible topic. y = 1, 2, 3, and 4 represents Topic 1, Topic 2, Topic 3, and Topic 4 correspondingly. Green marker shows the word that matches with any keyword of predicted topic. Red marker shows the mismatch. Best viewed in electronic format (zoomed in).}
    \label{fig:expl_4}
\end{figure}
\subsection{Comparison with Ground Truth}
To compare with ground truth, we extracted 500 tweets from train data and test data (new tweets) and did manual annotation both for train/test data based on functionality of tweets (described in \hyperref[subsec:4.3]{Subsection 4.3}). We achieved approximately 93\% train accuracy and 80\% test accuracy.

\subsection{Explanation Observation}
In Table \ref{tab:observ}, we show some observation of explanations of the twitter mining.  Top two rows (1$^{st}$ and 2$^{nd}$ row) of Table \ref{tab:observ} show the tweets, predicted topic, annotated topic, mean KL divergence, and score from train data. Rest of the rows (3$^{rd}$, 4$^{th}$, 5$^{th}$, and 6$^{th}$ row) of Table \ref{tab:observ} show the tweets from the test dataset.

For the tweets from train data (1$^{st}$ and 2$^{nd}$ row of Table \ref{tab:observ}),  predicted topic and the annotated topic is same. Topic 2 has been inferred by human and predicted by the model (1$^{st}$ row of Table \ref{tab:observ}). In the 1$^{st}$ row, we have good accuracy (100\%) with mean KL divergence 0.026. For the 2$^{nd}$ row of Table \ref{tab:observ}, Topic 3 has been inferred by human and predicted by the model with lower mean KL divergence (0.015) having 95\% accuracy.
\begin{figure}[htbp]
  \centering  
  \includegraphics[width= 0.5\textwidth]{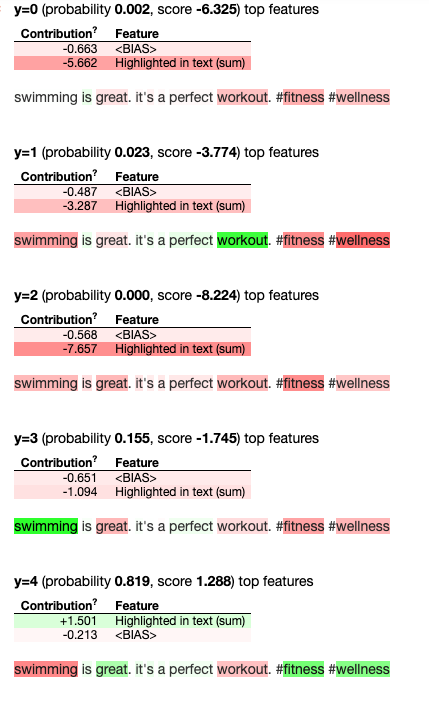}
    \caption{Explainable output of model prediction using Ex-Twit where human inference does not match with model prediction. Here, y represents topic number. y = 0 means incomprehensible topic. y = 1, 2, 3, and 4 represents Topic 1, Topic 2, Topic 3, and Topic 4 correspondingly. Green marker shows the word that matches with any keyword of predicted topic. Red marker shows the mismatch. Best viewed in electronic format (zoomed in).}
    \label{fig:expl_3}
\end{figure}
In Table \ref{tab:observ}, we show the tweets from test data where human annotation matches with model prediction in 3$^{rd}$, 4$^{th}$, and 5$^{th}$ row. Fig. \ref{fig:expl_4} shows the explainable output of model prediction for one of the tweets from test data (4$^{th}$ row of Table \ref{tab:observ}). This tweet is related to someone's weight loss journey by changing diet and going gym 5/6 times a week. We annotated this tweet with Topic 1 (7$^{th}$ column of Table \ref{tab:topic_keywords} shows the corresponding keywords).

In Fig. \ref{fig:expl_4}, y represents topic number and y = 0 means incomprehensible topic (we assign zero).  For y = 1, the words `diet' and `gym' (green marker) match with 2 topic keywords of Topic 1 (7$^{th}$ column of Table \ref{tab:topic_keywords}) containing the most frequent keyword (`diet'). For Topic 1, we achieved the highest prediction probability ($\approx$ 0.89) and the highest score ($\approx$ 2.4) in this tweet (4$^{th}$ row of Table \ref{tab:observ}). 

The 6$^{th}$ row of Table \ref{tab:observ}, we have tweets (having mean KL divergence = 0.061 and accuracy score = 92\%) from test data where the predicted topic and human annotated topic are not same. 

In Fig. \ref{fig:expl_3}, we show why the model prediction and manual annotation differs for the tweet of 6$^{th}$ row of Table \ref{tab:observ}. Human inferred the tweet as a swimming related topic (Topic 3). 9$^{th}$ column of Table \ref{tab:topic_keywords} shows the corresponding top-10 keywords of Topic 3. Our model predicted Topic 4 (10$^{th}$ column of Table \ref{tab:topic_keywords} shows the corresponding top-10 keywords) as the topic of the tweet. Ex-Twit can provide a possible explanation.

From Fig. \ref{fig:expl_3}, we observed that for y = 1, word `workout' (green marker) matches with one of the topic keywords of Topic 1 (7$^{th}$ column of Table \ref{tab:topic_keywords}). There are 4 mismatches (3 darker red markers and 1 lighter red marker). We can see the probability of prediction is 0.023. There is no matching keyword in Topic 2 for the tweet (when y = 2). For y = 3, word `swimming' (green marker) matches with the top frequent keyword of Topic 3 (9$^{th}$ column of Table \ref{tab:topic_keywords}) having higher prediction (prediction probability $\approx$ 0.16). This one also has 4 mismatches (red markers). It is noticeable that for y = 4, words  `fitness', `wellness', and `great' (green markers) match with 3 topic keywords of Topic 4 (10$^{th}$ column of Table \ref{tab:topic_keywords}) containing the top frequent keyword (`fitness'). Topic 4 achieved the highest prediction probability ($\approx$ 0.82) with the highest score ($\approx$ 1.3) among other topics for the tweet. Fig. \ref{fig:expl_3} shows the explanation of model prediction (Topic 4) for this tweet (last row of Table \ref{tab:observ}).

\section{Conclusions}
We proposed a method called Explainable Twitter Mining (Ex-Twit) combining LDA and LIME. We successfully showed the effectiveness of Ex-Twit in explanation of model predictions on Twitter health-related data. We believe that Ex-Twit can be applied more generally for the explanation of predictions in other NLP related applications. In the future, we would like to add more tweets to further validate our approach and observe the scalability. 

\bibliographystyle{named}
\bibliography{ijcai19.bib}

\end{document}